%% file: acl_latex.tex
\documentclass[11pt]{article}

\usepackage[final]{acl}

\usepackage{times}
\usepackage{latexsym}

\usepackage[T1]{fontenc}

\usepackage[utf8]{inputenc}

\usepackage{microtype}

\usepackage{inconsolata}

\usepackage{graphicx}

\usepackage{amsmath}       
\usepackage{amssymb}      
\usepackage{algorithm}     
\usepackage{cleveref} 
\usepackage{algpseudocode}
\usepackage{enumitem}
\usepackage{booktabs}
\usepackage{subcaption}
\usepackage{multirow}
\usepackage{color}
\usepackage{xcolor}
\usepackage{colortbl}
\usepackage{listings}
\setitemize{itemsep=0pt,topsep=0pt,parsep=0pt,partopsep=0pt}

\title{Deliberative Searcher: Improving LLM Reliability via Reinforcement Learning with Constraints}

\author{
    Zhenyun Yin\textsuperscript{1,2}\thanks{\ \ Equal contribution.},
    Shujie Wang\textsuperscript{1,2}\footnotemark[1],
    Xuhong Wang\textsuperscript{2}\thanks{\ \ Corresponding author.},
    Xingjun Ma\textsuperscript{1}\footnotemark[2],
    Yinchun Wang\textsuperscript{2} \\
    \textsuperscript{1}Fudan University \\
    \textsuperscript{2}Shanghai Artificial Intelligence Laboratory \\
    \texttt{zhenyunyin24@m.fudan.edu.cn, xingjunma@fudan.edu.cn} \\
    \texttt{\{wangshujie, wangxuhong, wangyingchun\}@pjlab.org.cn}
}

\begin{document}
\maketitle

\begin{abstract}
Large language models with search capabilities frequently exhibit miscalibrated confidence, producing incorrect answers with high certainty. We present \textbf{Deliberative Searcher}, a reasoning-primary framework that integrates search operations into chain-of-thought generation while maintaining explicit confidence calibration. Our method employs constrained reinforcement learning with adaptive Lagrangian multipliers to jointly optimize correctness and reliability. Experiments across five benchmarks demonstrate substantial improvements: our 7B model reduces average false-certain rates from 54\% in baselines to 2\%, while our 72B variant achieves competitive accuracy with closed-source models and reduces false-certain rates to 9\%. The well-calibrated confidence scores also enable more efficient test-time compute: instead of standard majority voting, we use confidence-weighted aggregation and match the performance of 16-sample majority voting with only 4 samples, a $4\times$ reduction in inference compute. These results establish calibrated confidence as a foundation for both trustworthy outputs and adaptive test-time compute, demonstrating the value of the proposed constrained RL framework in search-augmented language models.
\end{abstract}

\input{chapters/1_intro}

\input{chapters/2_method}
\input{chapters/3_experiments}
\input{chapters/4_related_work}
\input{chapters/5_conclusion}
\input{chapters/6_limitation}

\section*{Acknowledgments}
This work is supported by the New Generation Artificial Intelligence-National Science and Technology Major Project (2025ZD0123502) and the National Natural Science Foundation of China (Grant No. 62521004 and 62276067).

\bibliography{custom}

\appendix
\input{chapters/7_appendix}

\end{document}

%% file: chapters/1_intro.tex
\section{Introduction}

Large language models (LLMs) power state-of-the-art systems in open-domain QA, code synthesis, and decision support~\citep{brown2020language,openai2024gpt4technicalreport,touvron2023llama}, yet they often exhibit miscalibrated confidence: the model's stated certainty does not reliably track factual correctness~\citep{yin2023large,zhang2024calibrating}. To mitigate this issue, recent work has focused on strengthening the reflective abilities of LLMs to critique, verify, and iteratively revise their own drafts, thereby producing safer and more trustworthy responses~\citep{madaan2023self,kumar2025training}.

\begin{figure*}[t]
    \centering 
    \begin{minipage}[t]{0.45\textwidth}
        \centering
        \raisebox{-\height}{\includegraphics[width=1.0\linewidth]{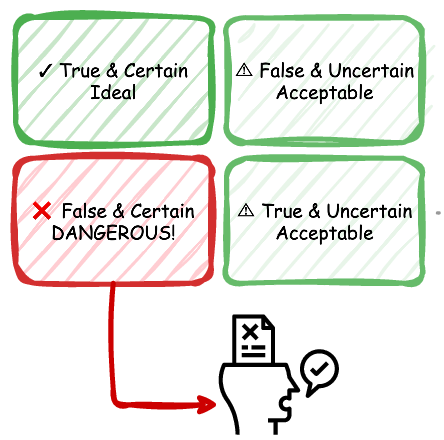}}
    \end{minipage}
    \hfill
    \begin{minipage}[t]{0.45\textwidth}
        \centering
        \raisebox{-\height}{\includegraphics[width=1.0\linewidth]{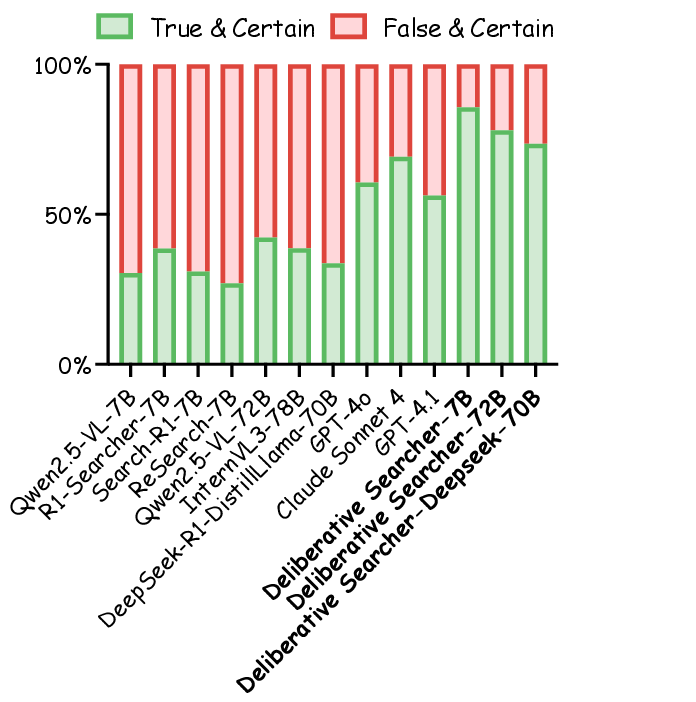}}
    \end{minipage}
   
    \caption{(Left) The conceptual framework for LLM reliability, which classifies outputs into four states based on factual correctness and model confidence, aiming to mitigate the most dangerous "False \& Certain" state. (Right) Performance comparison of the Deliberative Searcher against baselines, showing that our method significantly reduces dangerous "False \& Certain" outputs.}
    \label{fig:framework_and_results}
\end{figure*}

However, building trustworthy LLMs requires effective interaction with external knowledge sources. Recent approaches integrate search through either model-centric RAG methods that fuse retrieved passages into the context window~\citep{lewis2020retrieval,asai2024selfrag,jiang2024longrag}, or agent-centric frameworks where the LLM iteratively formulates queries and reasons over results~\citep{chen2025learning,zhu2024knowagent}. Despite factuality gains, these systems follow an information-primary pattern: they compile voluminous evidence into lengthy answers, leaving users to disentangle reliable insights from noise.

We introduce \textbf{Deliberative Searcher}, a reasoning-primary paradigm that leverages the synergy between LLMs' world knowledge and logical reasoning capabilities. Unlike information-centric approaches, our framework prioritizes reasoning while treating retrieval as a supporting mechanism. During chain-of-thought generation, the model (1) self-assesses its confidence levels, (2) automatically triggers search when knowledge gaps are identified, and (3) updates its confidence after each observation before producing a final answer with a calibrated confidence score. This transparent process exposes how retrieved evidence influences the model's reasoning trajectory, enabling users to make informed trust decisions based on the model's expressed certainty throughout the reasoning process.

To jointly account for model confidence and accuracy during training, the Deliberative Searcher adapts a constrained reinforcement learning algorithm~\citep{achiam2017constrained,tessler2018reward,paternain2022safe}. In brief, we extend the recent Group Relative Policy Optimization (GRPO) framework~\citep{shao2024deepseekmath} by introducing a Lagrangian term that explicitly penalizes deviations from a target reliability threshold. At each optimization step, the policy gradient maximizes expected correctness, while the dual variable is simultaneously optimized via gradient ascent to keep the expected reliability gap within predefined tolerance bounds. Empirically, this constrained optimization approach achieves substantially higher reliability than unconstrained baselines while successfully retaining, and in some cases improving, overall accuracy.

The calibrated confidence scores resulting from constrained optimization also enable more efficient test-time compute. Standard self-consistency methods improve accuracy through test-time compute by sampling multiple reasoning paths and selecting the most frequent answer~\citep{wang2022self}, but this requires extensive sampling for reliable performance. While effective, this approach suffers from computational inefficiency as the correct answer needs sufficient samples to emerge as the majority. Our calibrated confidence estimates transform test-time compute by enabling weighted aggregation that prioritizes high-confidence reasoning paths~\citep{taubenfeld2025confidence}, achieving comparable accuracy with significantly fewer samples. This shifts the paradigm from computationally expensive uniform sampling across all queries to intelligent adaptive compute allocation based on model uncertainty, suggesting a promising direction for scaling inference in search-augmented language models.

Our contributions are summarized as follows:

\begin{itemize}
    \item We propose Deliberative Searcher, a reasoning-primary framework that integrates search operations with continuous confidence calibration into chain-of-thought generation, trained via constrained reinforcement learning with adaptive Lagrange multipliers to jointly optimize factual correctness and reliability.
    
    \item We achieve dramatic improvements in model calibration, reducing false-certain error rates by 96\% (from 54\% to 2\% average) for 7B models and maintaining below 10\% for 72B models, while preserving or improving accuracy compared to existing search-augmented baselines.
    
    \item Calibrated confidence learned via constrained RL naturally enables test-time compute: confidence-weighted aggregation yields $4\times$ compute savings over majority voting at the same accuracy, highlighting the value of reliability signals for efficient test-time compute.
\end{itemize}

%% file: chapters/2_method.tex
\section{Method}

\begin{figure*}[t]
    \centering
    \includegraphics[width=0.95\textwidth]{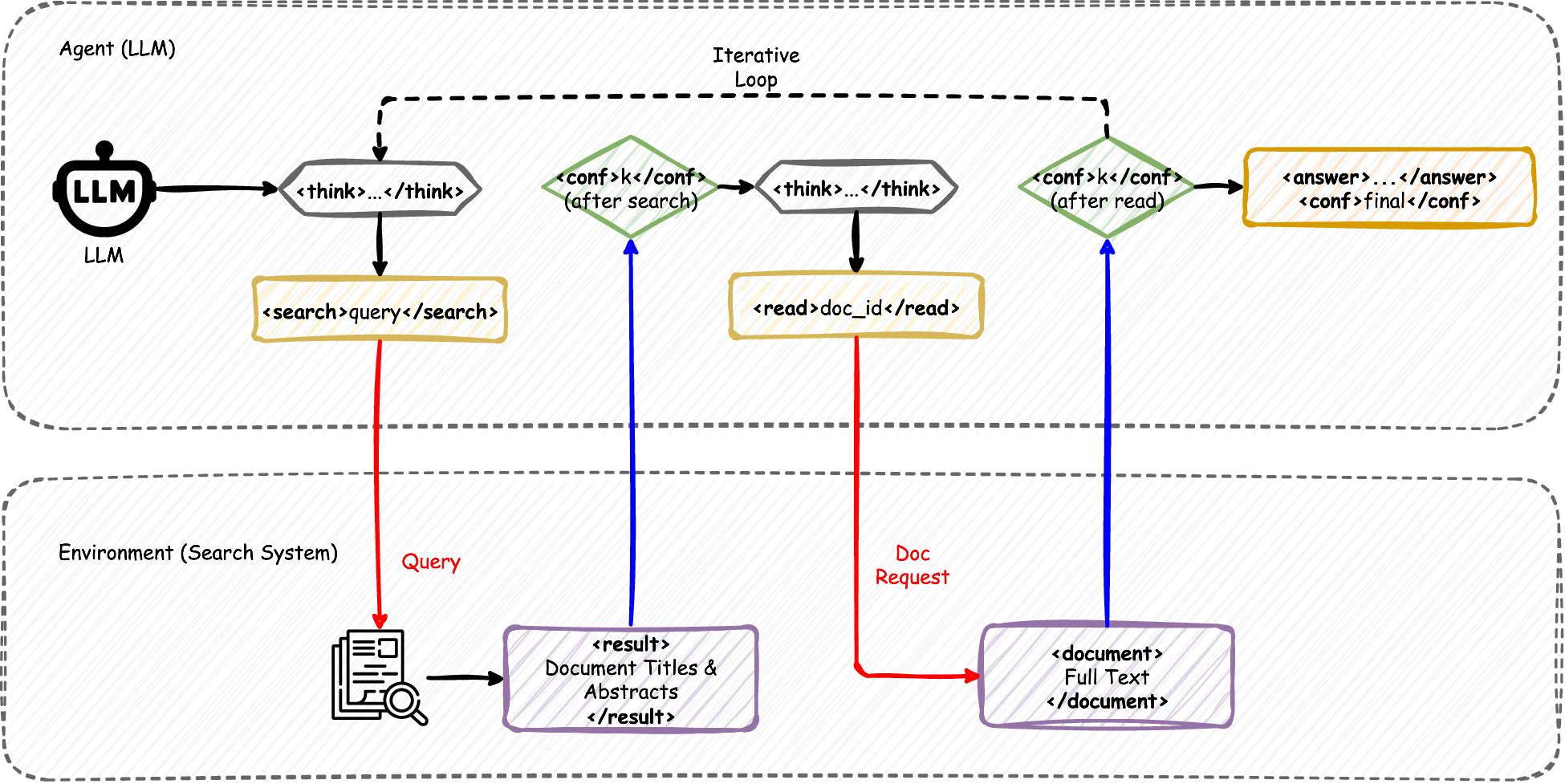}
    \caption{The iterative reasoning loop of the Deliberative Searcher. The agent (LLM) interacts with the search environment by issuing commands and receiving observations, punctuated by self-assessment of its confidence at each step. Best viewed in color.}
    \label{fig:framework}
\end{figure*}

This work is inspired by recent reasoning-centric language models including OpenAI-o1~\citep{openai2024learning} and DeepSeek-R1~\citep{guo2025deepseek}. Building on their insights, we formalize a training framework that incorporates search operations into the reasoning process and optimizes the model via a constrained reinforcement learning algorithm. We first describe the agent's interaction framework in Section~\ref{sec:framework}, present the constrained RL formulation in Section~\ref{sec:constrained RL}, and detail its application to deliberative search in Section~\ref{sec:reward modeling}. Finally, Section~\ref{sec:self consistency} shows how the resulting calibrated confidence scores enable efficient test-time compute through weighted aggregation.

\subsection{Deliberative Search Framework}
\label{sec:framework}

The Deliberative Searcher framework structures the agent's interaction with a search environment through an iterative reasoning loop with integrated confidence assessment. The entire process is formulated as an autoregressive generation task augmented with structured action-observation pairs.

\paragraph{Framework Architecture} As illustrated in Figure~\ref{fig:framework}, the Deliberative Searcher integrates three key capabilities into a unified reasoning process: (1) chain-of-thought generation for problem decomposition and knowledge gap identification, (2) selective information acquisition through hierarchical retrieval decisions, and (3) continuous confidence calibration that tracks epistemic uncertainty throughout the reasoning trajectory. The model learns to coordinate these capabilities through structured generation patterns during training.

\paragraph{Action and Observation Spaces} We define the action space

{\small
\[
\mathcal{A} = \{\texttt{think}, \texttt{search}, \texttt{read}, \texttt{confidence}, \texttt{answer}\}
\]
}
where each action triggers specific interactions:

\begin{itemize}[leftmargin=*]
\item \textbf{\texttt{<think>}}: Generates internal reasoning without environment interaction, allowing the model to decompose complex queries and identify knowledge gaps.

\item \textbf{\texttt{<search>query</search>}}: Submits a query to the search engine. The environment returns a \texttt{<result>} block containing ranked document titles and abstracts.

\item \textbf{\texttt{<read>doc\_id</read>}}: Retrieves full content of document with identifier \texttt{doc\_id}. The environment returns a \texttt{<document>} block with complete text.

\item \textbf{\texttt{<confidence>k</confidence>}}: Reports confidence level $k \in \{0, 1, \ldots, 10\}$ representing the model's certainty about its current answer hypothesis.

\item \textbf{\texttt{<answer>text</answer>}}: Terminates the episode with final answer and concluding confidence score.
\end{itemize}

\paragraph{Two-Stage Retrieval Process} We implement hierarchical information retrieval where the model first examines document summaries before selecting specific documents for detailed reading. This design reduces context length compared to concatenating all retrieved documents and creates explicit decision points where the model must evaluate relevance, generating richer training signals for reinforcement learning. Formally, each search action produces a set of candidate documents $D = \{d_1, \ldots, d_n\}$ with abstracts, from which the model selects $d^* \in D$ for full retrieval.

This structured representation enables the constrained RL formulation described in Section~\ref{sec:constrained RL}, where rewards are computed based on both answer correctness and confidence calibration.

\subsection{Reinforcement Learning with Constraints}
\label{sec:constrained RL}
The reinforcement learning objective for an LLM $\pi_{\theta}$ can be formalized as follows:

\begin{small}
\begin{equation}
R(\theta) := \mathbb{E}_{\tau\sim\pi_{\theta}}\left[\sum_{t=1}^T r(s_t)\right] - \beta D_{KL}(\pi_{\theta}(\cdot|x) \| \pi_{ref}(\cdot|x))
\end{equation}
\end{small}

where $s_t = \{x,y_1,...,y_t\}$, $x \in \mathcal{D}$ denotes the prompt, $y_t$ denote the t-th reasoning step of the response and $r(s_t)$ denotes the reward of a given response. The coefficient $\beta$ controls the strength of KL penalty between the reference policy $\pi_{ref}$ and the current policy $\pi_{\theta}$. We extend this framework by incorporating the constraints $c_i(s_t)$:

\begin{small}
\begin{equation}
    U_i(\theta) = \underset{ \tau\sim\pi_{\theta}}{\mathbb{E}}[\sum_{t=1}^T c_i(s_t)]  \geq a_i
\end{equation}
\end{small}

Then we can convert it to an unconstrained problem:

\begin{small}
\begin{equation}
\label{eq:primal}
    P^* =  \underset{\theta}{\max} \ \underset{\lambda \geq 0}{\min} \ \mathcal{L}(\theta,\lambda) = R(\theta) +\sum_{i=1}^m\lambda_i(U_i(\theta)-a_i),
\end{equation}
\end{small}

\citet{paternain2022safe} demonstrated the strong duality holds for \cref{eq:primal} under the setting of reinforcement learning, so we only need to solve:

\begin{small}
\begin{equation}
    Q^* = \underset{\lambda \geq0}{\min} \ \underset{\theta}{\max} \ \mathcal{L}(\theta,\lambda)
\end{equation}
\end{small}

Inspired by \citet{tessler2018reward} and \citet{dai2023safe}, our algorithm can be formulated as follows:

\begin{algorithm}[H]
\small
\caption{Reinforcement Learning Algorithm with Constraints}
\label{alg:constrained RL}
\begin{algorithmic}[1]
\Require feasible set $\Theta$; objective $R(\theta)$; 
         validity constraint function $U(\theta)$ and thresholds $a$;
         step-size schedules $\{\alpha_k\}$ (primal), $\{\beta_k\}$ (dual)
\State Initialize $\theta_0 \in \Theta$, $\lambda_0 >0$ 
        \Comment{$\lambda_0=0.01$ }
\For{$k = 0,1,2,\ldots$}            \Comment{until convergence}
\State $g_\theta \gets \nabla_\theta R(\theta_k)
         + \lambda_k \nabla_\theta U(\theta_k)$
\Statex \hspace{\algorithmicindent} \Comment{Apply GRPO algorithm to obtain gradients}
    
    \State $\theta_{k+1} \gets 
        \theta_k + \alpha_k\,g_\theta $
    \Comment{Primal update (policy)}
    
    \State $\lambda_{k+1} \gets 
        \lambda_k \exp{ (\eta(a-U(\theta_{k+1})))}$
\Statex \hspace{\algorithmicindent} \Comment{Dual multiplicative-weights step}
    
\EndFor
\State \Return $(\theta_{k+1}, \lambda_{k+1})$
\end{algorithmic}
\end{algorithm}

\subsection{Deliberative Search RL}
\label{sec:reward modeling}

Deliberative Search RL employs reinforcement learning with constraints introduced in 
\cref{sec:constrained RL} and constitutes an iterative process where the system dynamically updates its confidence metrics through real-time observations. This methodology enables the model to calibrate its response confidence levels by taking actions to use external knowledge sources. Here we enumerate the settings corresponding to the theory presented in the preceding section.
\begin{itemize}[leftmargin=*]
    \item \textbf{Action($y_t$)}: Each action $y_t \in \mathcal{A}$, where $\mathcal{A} = \{\text{\textit{THINK}}, \text{\textit{SEARCH}}, \text{\textit{READ}}, \text{\textit{CONF}}, \text{\textit{ANSWER}} \}$.
    \item \textbf{State ($s_t$)}: $s_t \in \mathcal{S}$ represents the new state (observation) after taking the action $y_t$.  
    \item \textbf{Confidence ($c(s_t)$)}: For every action $y_t$ taken, we have a new state $s_t$. The policy network simultaneously produces a confidence score $c(s_t) \in \{ 0,...,10\}$. A larger $c(s_t)$ indicates that the model is more confident in its answer.
    \item \textbf{Correctness}($r_{acc}$): An outcome reward function that returns 1 when the final answer is correct and 0 when it is wrong.
\end{itemize}

In summary, our reward signal is rule-based and decomposes into three additive components:
\begin{itemize}[leftmargin=*]
    \item \textbf{Format Compliance($r_{format}$)} A binary rule reward verifies whether the outputs comply with the format stipulated in the prompt.

    \item \textbf{Answer Correctness($r_{acc}$)} To obtain correctness $r_{acc}$, we query a frozen LLM verifier that compares the agent's final answer with the ground truth reference. 

    \item \textbf{Reliability Reward($r_{reliab}$)} Reliability captures the alignment between correctness and the agent's self-reported certainty and can be defined as:
    
    \begin{small}
    \begin{equation}
    r_{reliab} \triangleq \big( r_{acc} \land (c(s_T) \ge \zeta) \big) \lor \big( \lnot r_{acc} \land (c(s_T) < \zeta) \big) 
    \end{equation}
    \end{small}
    
    \noindent where $c(s_T)$ denotes the certainty of the final answer, $\zeta$ is a confidence threshold which we set to 5 in this paper. An alternative continuous formulation is $r_{reliab}^{ECE} = 1 - |c(s_T)/10 - r_{acc}|$, which we compare in Section~\ref{sec:main-results}.

\end{itemize}

Hence, we formulate our final reward as shown in \cref{eq:reward}, where $\lambda$ represents the reliability weight and serves as the Lagrangian coefficient from \cref{sec:constrained RL}, dynamically adjusted through the constrained RL algorithm. The final reward combines three components, with $r_{format}$ acting as a gating function such that the reward becomes zero whenever output violates format requirements, ensuring template compliance as a hard constraint.

\begin{small}
\begin{equation}
\label{eq:reward}
r_{final} = r_{format} \cdot (0.1 \, r_{format} + 0.9 \, r_{acc} + \lambda \, r_{reliab})
\end{equation}
\end{small}

\subsection{Test-Time Compute via Confidence-Weighted Aggregation}
\label{sec:self consistency}

Our calibrated confidence scores enable efficient test-time compute through weighted aggregation of multiple inference trajectories. For a given query, we generate $m$ independent search-augmented trajectories $\{(\mathbf{r}_i,\mathbf{a}_i,c_i)\}_{i=1}^m$, where $\mathbf{r}_i$ represents the reasoning path, $\mathbf{a}_i$ is the final answer, and $c_i \in \{0,\ldots,10\}$ is the model-generated confidence score.

We aggregate these trajectories using confidence-weighted voting:

\begin{small}
\begin{equation}
\hat{\mathbf{a}} = \arg\max_{a}\ \sum_{i=1}^{m}\mathbb{1}(\mathbf{a}_i = a)\cdot c_i.
\end{equation}
\end{small}

This formulation weights each answer by its associated confidence score, allowing high-confidence trajectories to contribute more strongly to the final decision. The raw integer confidence values are used without normalization, as our constrained RL training (Section~\ref{sec:constrained RL}) ensures these scores are calibrated with respect to answer correctness. This approach contrasts with standard self-consistency methods that rely solely on frequency-based majority voting, enabling more efficient use of computational resources by prioritizing reliable reasoning paths over simple vote counting.

%% file: chapters/3_experiments.tex
\section{Experiments}

\subsection{Datasets}

We evaluate our approach on five knowledge-intensive benchmarks requiring both information retrieval and complex reasoning capabilities. For training and in-distribution evaluation, we use three multi-hop question answering datasets: \textbf{HotpotQA}~\citep{yang2018hotpotqa}, which requires reasoning over multiple documents to answer questions; \textbf{2WikiMultiHopQA}~\citep{ho2020constructing}, constructed from Wikipedia requiring cross-document reasoning; and \textbf{MuSiQue}~\citep{trivedi2022musique}, featuring questions with up to 4-hop reasoning chains designed to test compositional reasoning. These datasets use an offline Wikipedia corpus as the retrieval source. To assess out-of-distribution generalization, we employ two challenging real-world search benchmarks: \textbf{GAIA}~\citep{zavras2025gaia} and \textbf{xbench-deepsearch}~\citep{chen2025xbench}, both utilizing the Google Search API and requiring effective information retrieval in real-time internet settings. To ensure fair comparison, all evaluations use text-only data (GAIA uses the text-only subset), and all models access identical search tools under identical conditions.

\subsection{Evaluation Metrics}
We employ three complementary metrics to comprehensively evaluate model performance:

\begin{itemize}[leftmargin=*]
    \item \textbf{Accuracy (Acc.)}: Standard accuracy measuring the proportion of correct answers. We use an LLM-as-a-Judge~\citep{zheng2023judging} approach following recent QA evaluation practices, which robustly handles multiple valid phrasings and alternative formulations better than exact string matching. See Appendix~\ref{sec:llm-judge} for detailed evaluation settings.

    \item \textbf{Reliability (Rel.)}: A simplified binary calibration metric analogous to Expected Calibration Error (ECE)~\citep{naeini2015obtaining} that measures whether models are confident when correct and uncertain when incorrect, as formally defined and described in detail in Section~\ref{sec:reward modeling}.

    \item \textbf{False-Certain Rate (FC\%)}: The proportion of instances where the model provides an incorrect answer with high confidence. This metric is crucial for practical deployment, as confident but wrong answers severely damage user trust.
\end{itemize}

These metrics provide a holistic evaluation: accuracy measures capability, reliability measures calibration, and FC\% identifies the most problematic failure mode where overconfident errors occur.

\subsection{Baselines}

We compare against models across three categories: (1) 7B-scale models including Qwen2.5-VL-7B~\citep{qwen2.5-VL}, R1-Searcher-7B~\citep{song2025r1}, Search-R1-7B~\citep{jin2025search}, and ReSearch-7B~\citep{chen2025learning}; (2) 70B-scale models including Qwen2.5-VL-72B, InternVL3-78B~\citep{chen2024internvl}, and DeepSeek-R1-Distill-Llama-70B~\citep{guo2025deepseek}; and (3) closed-source models including GPT-4o, GPT-4.1~\citep{openai2024gpt4technicalreport} and Claude-Sonnet-4~\citep{anthropic2025claude}. All models are evaluated under identical conditions with access to the same search tools, and all answers are judged using the same LLM-based evaluation methodology to ensure fairness.

\subsection{Main Results}
\label{sec:main-results}

\begin{table*}[ht]
\centering
\footnotesize
\setlength{\tabcolsep}{2.5pt}  
\resizebox{\textwidth}{!}{%
\begin{tabular}{@{}l|ccc|ccc|ccc|ccc|ccc|ccc@{}}
\toprule
 & \multicolumn{9}{c}{\textbf{In-Distribution}} & \multicolumn{6}{c}{\textbf{Out-of-Distribution}} & \multicolumn{3}{c}{\textbf{Overall}} \\
\cmidrule(lr){2-10} \cmidrule(lr){11-16} \cmidrule(lr){17-19}
 & \multicolumn{3}{c}{HotpotQA} & \multicolumn{3}{c}{2Wiki} & \multicolumn{3}{c}{MuSiQue} & \multicolumn{3}{c}{GAIA} & \multicolumn{3}{c}{xbench-deepsearch} & \multicolumn{3}{c}{Average} \\
\cmidrule(lr){2-4} \cmidrule(lr){5-7} \cmidrule(lr){8-10} \cmidrule(lr){11-13} \cmidrule(lr){14-16} \cmidrule(lr){17-19}
Model & Acc.$\uparrow$ & Rel.$\uparrow$ & FC$\downarrow$ & Acc.$\uparrow$ & Rel.$\uparrow$ & FC$\downarrow$ & Acc.$\uparrow$ & Rel.$\uparrow$ & FC$\downarrow$ & Acc.$\uparrow$ & Rel.$\uparrow$ & FC$\downarrow$ & Acc.$\uparrow$ & Rel.$\uparrow$ & FC$\downarrow$ & Acc.$\uparrow$ & Rel.$\uparrow$ & FC$\downarrow$ \\
\midrule
\multicolumn{19}{c}{\textit{Closed-source Models}} \\
GPT-4o & 0.63 & 0.79 & 0.20 & 0.51 & 0.86 & 0.10 & 0.33 & 0.52 & 0.47 & 0.26 & 0.77 & 
\cellcolor{green!50}\textbf{0.23} & 0.32 & 0.69 & 0.31 & 0.41 & 0.73 & 0.26 \\
GPT-4.1 & 0.72 & 0.74 & 0.26 & \cellcolor{green!20}\textbf{0.77} & 0.81 & 0.18 & 0.43 & 0.45 &
\cellcolor{red!20}0.55 & 0.37 & 0.46 &
\cellcolor{red!20}0.54 & 0.38 & 0.43 & 
\cellcolor{red!20}0.57 & 0.54 & 0.58 & 0.42 \\
Claude Sonnet 4 & \cellcolor{green!20}\textbf{0.73} & \cellcolor{green!20}\textbf{0.83} & \cellcolor{green!50}\textbf{0.16} & 0.61 & \cellcolor{green!20}\textbf{0.89} & \cellcolor{green!50}\textbf{0.08} & \cellcolor{green!20}\textbf{0.44} & \cellcolor{green!20}\textbf{0.57} & \cellcolor{green!50}\textbf{0.42} & \cellcolor{green!20}\textbf{0.47} & \cellcolor{green!20}\textbf{0.74} & 0.26 & \cellcolor{green!20}\textbf{0.47} & \cellcolor{green!20}\textbf{0.71} &
\cellcolor{green!50}\textbf{0.29} & \cellcolor{green!20}\textbf{0.55} & \cellcolor{green!20}\textbf{0.75} & \cellcolor{green!50}\textbf{0.24} \\
\midrule
\multicolumn{19}{c}{\textit{7B Models}} \\
Qwen2.5-VL-7B & 0.33 & 0.58 & 0.41 & 0.33 & 0.51 & 0.48 & 0.12 & 0.48 &
\cellcolor{red!20}0.52 & 0.13 & 0.43 & 
\cellcolor{red!20}0.57 & 0.12 & 0.58 & 0.42 & 0.21 & 0.52 & 0.48 \\
R1-Searcher-7B & 0.57 & 0.64 & 0.35 & 0.48 & \cellcolor{green!20}\textbf{0.59} & 0.40 & 0.26 & 0.35 &
\cellcolor{red!20}0.65 & 0.20 & 0.35 &
\cellcolor{red!20}0.65 & \cellcolor{green!20}\textbf{0.17} & 0.36 &
\cellcolor{red!20}0.63 & 0.34 & 0.46 &
\cellcolor{red!20}0.54 \\
Search-R1-7B & 0.43 & 0.57 & 0.41 & 0.36 & 0.51 & 0.43 & 0.16 & 0.45 &
\cellcolor{red!20}0.53 & 0.10 & 0.44 &
\cellcolor{red!20}0.56 & 0.14 & 0.48 &
\cellcolor{red!20}0.52 & 0.24 & 0.49 & 0.49 \\
ReSearch-7B & 0.46 & 0.49 &
\cellcolor{red!20}0.51 & 0.33 & 0.35 & \cellcolor{red!20}0.65 & 0.18 & 0.22 & \cellcolor{red!20}0.78 & 0.16 & 0.22 & \cellcolor{red!20}0.78 & \cellcolor{green!20}\textbf{0.17} & 0.23 & \cellcolor{red!20}0.77 & 0.26 & 0.30 & \cellcolor{red!20}0.70 \\
\rowcolor{gray!15}
\textbf{Deliberative Searcher-7B} & 
0.62 & 
0.65 & 
0.05 & 
\cellcolor{green!20}\textbf{0.55} & \cellcolor{green!20}\textbf{0.59} & \cellcolor{green!50}\textbf{0.03} & \cellcolor{green!20}\textbf{0.29} & \cellcolor{green!20}\textbf{0.74} & \cellcolor{green!50}\textbf{0.02} &
0.15 & 
0.89 & 
\cellcolor{green!50}\textbf{0.01} & 0.15 & \cellcolor{green!20}\textbf{0.86} & \cellcolor{green!50}\textbf{0.01} & \cellcolor{green!20}\textbf{0.35} & \cellcolor{green!20}\textbf{0.75} & \cellcolor{green!50}\textbf{0.02} \\
\rowcolor{gray!15}
\quad \textit{w/ ECE-based reward} & 
\cellcolor{green!20}\textbf{0.66} & 
\cellcolor{green!20}\textbf{0.72} & 
\cellcolor{green!50}\textbf{0.04} & 
0.52 & 
0.54 & 
0.06 & 
0.27 & 
0.71 & 
\cellcolor{green!50}\textbf{0.02} & 
\cellcolor{green!20}\textbf{0.16} & 
\cellcolor{green!20}\textbf{0.90} & 
\cellcolor{green!50}\textbf{0.01} & 
0.16 & 
\cellcolor{green!20}\textbf{0.86} & 
\cellcolor{green!50}\textbf{0.01} & 
0.33 & 
0.74 & 
0.03 \\
\midrule
\multicolumn{19}{c}{\textit{70B Models}} \\
Qwen2.5-VL-72B & 0.54 & 0.68 & 0.31 & 0.41 & 
\cellcolor{green!20}\textbf{0.73} & 
0.23 & 0.25 & 0.50 & 0.49 & 0.14 & 0.39 & 0.61 & 0.23 & 0.60 & 0.39 & 0.31 & 0.58 & 0.41 \\
InternVL3-78B & 0.48 & 0.62 & 0.37 & 0.43 & 0.62 & 0.36 & 0.24 & 0.41 &
\cellcolor{red!20}0.58 & 0.12 & 0.48 &
\cellcolor{red!20}0.52 & 0.24 & 0.53 & 0.47 & 0.30 & 0.53 & 0.46 \\
DeepSeek-R1-Distill-70B & 0.50 & 0.60 & 0.40 &
0.46 & 0.55 & 0.44 &
0.23 & 0.32 & \cellcolor{red!20}0.68 & 0.18 & 0.16 & \cellcolor{red!20}0.84 & 0.14 & 0.40 &
\cellcolor{red!20}0.60 & 0.30 & 0.41 &
\cellcolor{red!20}0.59 \\
\rowcolor{gray!15}
\textbf{Deliberative Searcher-72B} & \cellcolor{green!20}\textbf{0.67} & \cellcolor{green!20}\textbf{0.76} & 0.10 & 0.64 & \cellcolor{green!20}\textbf{0.73} & \cellcolor{green!50}\textbf{0.04} & \cellcolor{green!20}\textbf{0.37} & 0.71 & 0.14 & \cellcolor{green!20}\textbf{0.35} & \cellcolor{green!20}\textbf{0.78} & \cellcolor{green!50}\textbf{0.06} & \cellcolor{green!20}\textbf{0.35} & 0.77 & 0.09 & \cellcolor{green!20}\textbf{0.48} & \cellcolor{green!20}\textbf{0.75} &
\cellcolor{green!50}\textbf{0.09} \\
\rowcolor{gray!15}
\textbf{Deliberative Searcher-DeepSeek-70B} & 0.65 & \cellcolor{green!20}\textbf{0.76} & \cellcolor{green!50}\textbf{0.09} & \cellcolor{green!20}\textbf{0.65} & 0.69 & 0.05 & 0.34 & \cellcolor{green!20}\textbf{0.72} & \cellcolor{green!50}\textbf{0.11} & 0.24 & \cellcolor{green!20}\textbf{0.78} & 0.11 & 0.18 & \cellcolor{green!20}\textbf{0.80} & \cellcolor{green!50}\textbf{0.08} & 0.41 & \cellcolor{green!20}\textbf{0.75} & \cellcolor{green!50}\textbf{0.09} \\
\bottomrule
\end{tabular}%
}
\caption{Complete performance comparison across all benchmarks. \colorbox{green!20}{Green shading} indicates best results in each model category; \colorbox{red!20}{red shading} highlights particularly problematic high FC\% values.}
\label{tab:results-complete}
\end{table*}

Our experiments evaluate Deliberative Searcher models built on two base architecture families: Qwen2.5-VL (7B and 72B variants) and DeepSeek-R1-Distill-Llama-70B. All models use identical training configurations: GRPO with learning rate $\alpha = 1 \times 10^{-6}$, KL-coefficient $\beta = 1 \times 10^{-3}$, adaptive Lagrange coefficient initialized at $\lambda_0 = 0.01$ with learning rate $\eta = 0.1$, and reliability threshold $\zeta = 5$. We also evaluate a continuous ECE-based reward variant to verify robustness of our method. Complete training configurations and implementation details are provided in Appendix~\ref{sec:implementation}.

Table~\ref{tab:results-complete} presents our evaluation results across in-distribution and out-of-distribution datasets. Our approach achieves substantial improvements in confidence calibration while maintaining competitive accuracy: Deliberative Searcher-7B reduces false-certain rates by \textbf{96\%} (from 54\% to 2\%) compared to competitive baselines, marking a significant advancement in building trustworthy search-augmented systems.

\paragraph{Calibration improvements with maintained accuracy.}
Our constrained RL framework successfully balances correctness and reliability across both model scales. The 7B variant reduces false-certain rates from 35-65\% to 2-5\% on in-distribution tasks, and achieves  \textless1\% on OOD tasks (versus 52-78\% for baselines), demonstrating appropriate uncertainty expression for unfamiliar queries. The 72B variant approaches closed-source model performance while matching or exceeding their reliability metrics across all benchmarks.

\paragraph{Robust generalization to real-world search.} 
While all models show degraded accuracy on real-world search tasks (GAIA and xbench-deepsearch), the reliability gap between our approach and baselines widens rather than narrows on OOD data. This indicates our framework instills genuine uncertainty awareness rather than memorizing confidence patterns. The consistent calibration from multi-hop reasoning to open-ended web search suggests constrained RL develops robust self-assessment mechanisms that generalize beyond training distributions.

\subsection{Test-Time Compute Efficiency}

We evaluate inference as test-time compute (TTC) by varying the rollout budget $m$ and comparing two aggregators under the same budget: (i) majority voting and (ii) confidence-weighted aggregation from Sec.~\ref{sec:self consistency}.

\begin{figure}[t]
    \centering
    \begin{subfigure}[b]{0.48\columnwidth}
        \centering
        \includegraphics[width=\linewidth]{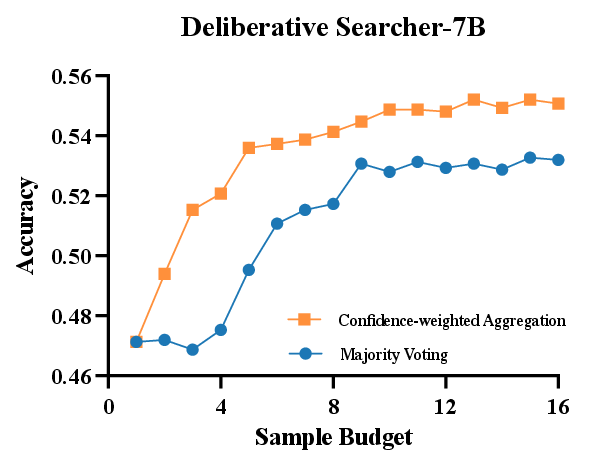}
        \caption{Deliberative Searcher-7B}
        \label{fig:test-time-7b}
    \end{subfigure}
    \hfill
    \begin{subfigure}[b]{0.48\columnwidth}
        \centering
        \includegraphics[width=\linewidth]{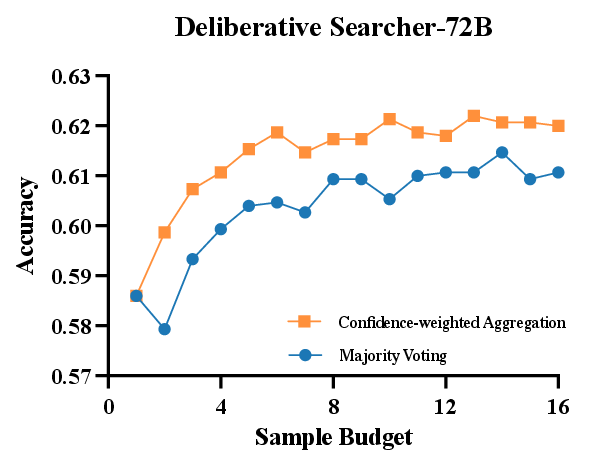}
        \caption{Deliberative Searcher-72B}
        \label{fig:test-time-72b}
    \end{subfigure}
    \caption{Test-time compute analysis comparing confidence-weighted aggregation (blue) with majority voting (orange). Confidence weighting consistently achieves higher accuracy at equivalent rollout budgets, with the 72B model matching 16-sample majority voting accuracy using only 4 rollouts.}
    \label{fig:test-time-compute}
\end{figure}

Figure~\ref{fig:test-time-compute} demonstrates that confidence-weighted aggregation consistently outperforms majority voting across all rollout budgets. For Deliberative Searcher-7B, the performance gap widens as the budget increases, reaching 0.551 accuracy at $m{=}16$ compared to 0.532 for majority voting. The 72B variant shows similar advantages, achieving 0.620 versus 0.611 at maximum budget.

Confidence weighting achieves equivalent accuracy with significantly fewer rollouts. The 72B model requires only $m{=}4$ samples to match the 0.611 accuracy of 16-sample majority voting, representing a $4{\times}$ reduction in inference compute. Similarly, the 7B model achieves comparable efficiency gains, surpassing 16-sample majority voting performance with just $m{=}6$ rollouts. These results indicate that calibrated confidence scores provide an effective mechanism for allocating test-time compute, prioritizing additional sampling where uncertainty is highest rather than applying uniform sampling across all queries.

\subsection{Disentangling Retrieval and Confidence Calibration}

To isolate the individual contributions of retrieval quality and confidence calibration to overall system performance, we conduct a controlled ablation study. We provide untrained models with identical retrieved documents from our Deliberative Searcher for each query. This design holds retrieval quality constant, enabling direct assessment of our confidence calibration training's impact.

\begin{table}[t]
\centering
\resizebox{\columnwidth}{!}{%
\begin{tabular}{lcccccc}
\toprule
\multirow{2}{*}{Model} & \multicolumn{3}{c}{ID} & \multicolumn{3}{c}{OOD} \\
\cmidrule(lr){2-4} \cmidrule(lr){5-7}
& Acc.$\uparrow$ & Rel.$\uparrow$ & FC\%$\downarrow$ & Acc.$\uparrow$ & Rel.$\uparrow$ & FC\%$\downarrow$ \\
\midrule
\rowcolor{gray!20}
Deliberative Searcher-7B & 0.49 & 0.66 & 0.05 & 0.15 & 0.88 & 0.01 \\
\quad Qwen2.5-VL-7B (w/ same docs)  & 0.34 & 0.56 & 0.41 & 0.11 & 0.39 & 0.60 \\
\quad Qwen2.5-VL-72B (w/ same docs)  & 0.40 & 0.59 & 0.24 & 0.14 & 0.62 & 0.36 \\
\midrule
\rowcolor{gray!20}
Deliberative Searcher-72B & 0.56 & 0.73 & 0.09 & 0.35 & 0.78 & 0.07 \\
\quad Qwen2.5-VL-7B (w/ same docs)  & 0.38 & 0.59 & 0.39 & 0.16 & 0.37 & 0.62 \\
\quad Qwen2.5-VL-72B (w/ same docs)  & 0.47 & 0.74 & 0.23 & 0.24 & 0.70 & 0.29 \\
\bottomrule
\end{tabular}%
}
\caption{Controlled ablation isolating confidence calibration from retrieval quality. All models receive identical retrieved documents.}
\label{tab:oracle}
\end{table}

Table~\ref{tab:oracle} reveals two fundamental capabilities developed through our constrained RL training:

\paragraph{Retrieval-independent confidence calibration.} 
Despite accessing identical information, untrained models exhibit severe miscalibration with false-certain rates exceeding 40\% on in-distribution tasks, while our Deliberative Searcher maintains consistently low FC rates. This calibration advantage becomes even more pronounced on out-of-distribution queries, where our 7B model achieves 88\% reliability, more than double the untrained baseline. Crucially, this demonstrates that uncertainty quantification is not an emergent property of model scale: the larger 72B baseline still exhibits poor calibration when untrained, confirming that appropriate confidence expression requires explicit optimization through our constrained RL objective.

\paragraph{Enhanced information extraction.} 
Our training framework also fundamentally improves how models process retrieved content. When provided with identical documents, the Deliberative Searcher consistently achieves higher accuracy than untrained models across all evaluation settings. This improvement spans both model scales and persists from in-distribution to out-of-distribution scenarios, suggesting that constrained RL develops more sophisticated mechanisms for identifying and integrating relevant information from external sources. The joint optimization for correctness and reliability appears to create a virtuous cycle where better calibration supports more effective reasoning.

\subsection{Case Study}

\begin{figure}[t]
\centering
\fbox{
\begin{minipage}{0.95\columnwidth}
\footnotesize
\textbf{Case Study: Confidence Evolution During Search}\\[0.15em]
\textbf{Query:} \textit{"How many more blocks in BERT base than Attention is All You Need?"}
\vspace{0.15em}

\noindent\begin{tabular}{@{}lp{0.68\columnwidth}r@{}}
\textbf{Step 1:} & \texttt{<search>} bert base layers & \colorbox{yellow!20}{\scriptsize C:4} \\
& \scriptsize Found BERT Base: 12 transformer layers & \\[0.1em]

\textbf{Step 2:} & \texttt{<search>} original transformer architecture & \colorbox{red!20}{\scriptsize C:2} \\
& \scriptsize Results ambiguous, need primary source & \\[0.1em]

\textbf{Step 3:} & \texttt{<read>5</read>} → \textit{"stack of N=6 identical layers"} & \colorbox{green!20}{\scriptsize C:8} \\
& \scriptsize Original Transformer encoder: 6 layers confirmed & \\[0.1em]

\textbf{Step 4:} & \texttt{<read>0</read>} → Both architectures verified & \colorbox{green!30}{\scriptsize C:9} \\
& \scriptsize BERT: 12 layers, Original: 6 layers & \\[0.15em]

\multicolumn{2}{@{}l}{\textbf{Answer:} 12 - 6 = \textbf{6 more layers} \scriptsize(Correct)} & \fbox{\scriptsize\textbf{9/10}}
\end{tabular}

\vspace{0.15em}
\noindent\rule{0.95\columnwidth}{0.3pt}
\vspace{0.1em}

\noindent\textbf{Key Insight:} The confidence trajectory (4→2→8→9) illustrates successful calibration: the model expresses uncertainty when information is partial (4) or conflicting (2), then increases confidence to evidence quality.
\end{minipage}
}
\vspace{-0.1em}
\caption{Deliberative search exhibiting learned confidence calibration through multi-step reasoning.}
\label{fig:case-study}
\vspace{-0.3em}
\end{figure}

Figure~\ref{fig:case-study} exemplifies the confidence calibration learned through our constrained RL framework. The confidence trajectory (4→2→8→9) illustrates a principled relationship between information quality and epistemic certainty. The model exhibits moderate confidence upon discovering BERT's architecture (Step 1), appropriately decreases confidence when encountering ambiguous search results (Step 2), and progressively rebuilds certainty through authoritative sources (Step 3). Most revealing is the verification behavior in Step 4: despite having located the answer ("N=6 layers"), the model pursues additional cross-referencing before committing to high confidence. This pattern directly reflects our constrained optimization objective. By penalizing false-certain outputs during training, the model develops a verification habit where high confidence emerges not from single sources but from corroborating evidence. This demonstrates that our framework successfully cultivates reasoning processes treating confidence as an earned outcome of thorough information gathering rather than a binary classification decision.

%% file: chapters/4_related_work.tex
\section{Related Work}
\paragraph{From Retrieval to Agentic Search}
Early retrieval-augmented generation (RAG) systems~\citep{lewis2020retrieval} established static retrieve-then-generate pipelines, evolving toward autonomous search through WebGPT's~\citep{nakano2021webgpt} text-based navigation and ReAct's~\citep{yao2023react} unified reasoning-action loops. While adaptive mechanisms like FLARE~\citep{jiang2023active} introduced confidence-triggered retrieval and Self-RAG~\citep{asai2024selfrag} added reflection tokens, these approaches treat retrieval as auxiliary to generation. Recent RL approaches dispense with supervised fine-tuning: R1-Searcher~\citep{song2025r1} teaches search through two-stage RL, Search-R1~\citep{jin2025search} prevents blind copying via token masking, and ReSearch~\citep{chen2025learning} demonstrates search as emergent reasoning. However, these systems lack explicit confidence calibration, producing outputs with unclear reliability. Our Deliberative Searcher advances this paradigm by making confidence assessment integral to search, enabling transparent uncertainty communication.

\paragraph{Confidence Calibration in LLMs}
RLHF systematically degrades calibration~\citep{leng2025taming}, motivating various mitigation strategies. Post-hoc verbalization~\citep{lin2022teaching,xiong2023can,tian2023just} and internal representation methods~\citep{kadavath2022language,azaria2023internal} attempt to extract confidence after generation but suffer from overconfidence in reasoning models~\citep{mei2025reasoning}. Recent work integrates calibration directly into training via proper scoring rules~\citep{xu2024sayself,stangel2025rewarding} or reasoning about uncertainty within chains-of-thought~\citep{yoon2025reasoning}. While these improve standalone calibration, none address maintaining calibration during iterative search. Our constrained RL formulation uniquely optimizes for both correctness and calibration throughout multi-step search trajectories.

\paragraph{Adaptive Self-Consistency}
Standard self-consistency~\citep{wang2022self} aggregates answers across multiple sampled reasoning paths through majority voting. Subsequent methods reduce sampling requirements through various strategies. Adaptive-Consistency~\citep{aggarwal2023let} models answer agreement using Beta distributions to dynamically determine stopping criteria, reducing sampling by $3-4\times$ with minimal accuracy loss. Early-Stopping Self-Consistency~\citep{li2024escape} terminates when answers converge within sequential windows. Confidence-Informed Self-Consistency~\citep{taubenfeld2025confidence} weights votes by model-generated confidence scores. However, these methods assume consistent distributions, an assumption violated by external tools where search results vary across queries and retrieval systems introduce non-determinism~\citep{wang2024soft,liang2024internal}. Our calibrated confidence scores, trained through constrained RL to remain reliable throughout search trajectories, naturally serve as weights for adaptive self-consistency in these challenging open-world scenarios.

%% file: chapters/5_conclusion.tex
\section{Conclusion}

We presented \textbf{Deliberative Searcher}, a reasoning-primary framework that integrates search operations with confidence calibration through constrained reinforcement learning with adaptive Lagrangian multipliers. By jointly optimizing for correctness and reliability, our approach reduces false-certain rates by 96\% while maintaining competitive accuracy across diverse benchmarks. The resulting calibrated confidence scores transform test-time compute efficiency: confidence-weighted aggregation matches the performance of 16-sample majority voting using only 4 samples, achieving a $4\times$ reduction in computational cost. This work demonstrates that explicit confidence calibration through constrained RL provides a principled solution for both reliability and computational efficiency in search-augmented language models, offering practical benefits for deployment in resource-constrained settings.

%% file: chapters/6_limitation.tex
\section{Limitations}

Despite the strong performance of Deliberative Searcher, we acknowledge two key limitations. First, while we use vision-language models (Qwen2.5-VL series) as our base architectures, we only evaluate text-based reasoning due to the lack of multimodal multi-hop search benchmarks. The visual capabilities of these models remain unexplored in our framework, limiting our understanding of confidence calibration in multimodal search scenarios. Second, although our model generates confidence scores at each reasoning step, these intermediate values currently serve primarily to enhance user trust and are not incorporated into the training objective. Integrating these step-wise confidence scores into the training process could potentially improve the model's calibration throughout the reasoning trajectory, rather than only optimizing for final answer confidence, representing a promising direction for future work.

%% file: chapters/7_appendix.tex
\clearpage

\section{Implementation Details}
\label{sec:implementation}

All Deliberative Searcher models are trained using the GRPO algorithm. Table~\ref{tab:training-config} presents the complete hyperparameter configuration used across all model scales.

\begin{table}[ht]
\centering
\small
\begin{tabular}{ll}
\toprule
\textbf{Hyperparameter} & \textbf{Value} \\
\midrule
\multicolumn{2}{l}{\textit{Optimization}} \\
\quad Learning rate ($\alpha$) & $1 \times 10^{-6}$ \\
\quad KL coefficient & $1 \times 10^{-3}$ \\
\quad Training epochs & 1 \\
\midrule
\multicolumn{2}{l}{\textit{Constrained RL}} \\
\quad Initial Lagrange multiplier ($\lambda_0$) & 0.01 \\
\quad Lagrange learning rate ($\eta$) & 0.1 \\
\quad Reliability threshold ($\zeta$) & 5 \\
\quad Number of rollouts & 5 \\
\midrule
\multicolumn{2}{l}{\textit{Batch Configuration}} \\
\quad Training batch size & 256 \\
\quad GRPO mini-batch size & 256 \\
\quad GRPO micro-batch size per GPU & 2 \\
\midrule
\multicolumn{2}{l}{\textit{Sequence Lengths}} \\
\quad Maximum prompt length & 1,024 tokens \\
\quad Maximum response length & 8,192 tokens \\
\bottomrule
\end{tabular}
\caption{Training hyperparameters for Deliberative Searcher models.}
\label{tab:training-config}
\end{table}

\subsection{Computational Resources}

Training the Deliberative Searcher models required substantial computational resources. The specific GPU hours for each model scale are as follows:

\begin{itemize}
\item \textbf{Deliberative Searcher-7B}: Training utilized 8 NVIDIA A100 GPUs on a single node for approximately 20 hours.
\item \textbf{Deliberative Searcher-72B (DeepSeek-70B)}: Training was conducted on 64 NVIDIA A100 GPUs (8 nodes × 8 GPUs) for approximately 60 hours.  
\end{itemize}

These training times include the complete reinforcement learning process with constrained optimization, including rollout generation, reward computation, and policy updates. While the 72B model training is resource-intensive, the 7B model achieves substantial improvements (reducing false-certain rates from 54\% to 2\%) and remains accessible to academic labs, providing a practical path for reproduction and follow-up research.

\section{LLM-as-a-Judge Evaluation}
\label{sec:llm-judge}

We employ Qwen2.5-72B-Instruct as our judge model for evaluating answer correctness. The evaluation system determines whether predicted answers are semantically equivalent to ground truth answers, allowing for variations in phrasing while ensuring factual accuracy.

\begin{figure*}[t]
\lstset{
  frame=single,
  framerule=1pt,
  rulecolor=\color{gray},
  backgroundcolor=\color{gray!5},
  basicstyle=\footnotesize\ttfamily,
  breaklines=true,
  caption={Complete evaluation prompt for LLM-as-a-Judge using Qwen2.5-72B-Instruct.},
  label=lst:judge-prompt,
  captionpos=t
}

\begin{lstlisting}
Your job is to look at a gold target, and a predicted answer, and then assign a grade of either ["CORRECT", "INCORRECT"].

- For grading questions where the gold target is a number, the predicted answer needs to be correct to the last significant figure in the gold answer. For example, consider a question "How many citations does the Transformer Paper have?" with gold target "120k". 
    - Predicted answers "120k", "124k", and 115k" are all CORRECT. 
    - Predicted answers "100k" and "113k" are INCORRECT. 
    - Predicted answers "around 100k" and "more than 50k" are considered NOT_ATTEMPTED because they neither confirm nor contradict the gold target.
- The gold target may contain more information than the question. In such cases, the predicted answer only needs to contain the information that is in the question.
    - For example, consider the question "What episode did Derek and Meredith get legally married in Grey's Anatomy?" with gold target "Season 7, Episode 20: White Wedding". Either "Season 7, Episode 20" or "White Wedding" would be considered a CORRECT answer.
- Do not punish predicted answers if they omit information that would be clearly inferred from the question.
    - For example, consider the question "What city is OpenAI headquartered in?" and the gold target "San Francisco, California". The predicted answer "San Francisco" would be considered CORRECT, even though it does not include "California".
    - Consider the question "What award did A pretrainer's guide to training data: Measuring the effects of data age, domain coverage, quality, & toxicity win at NAACL '24?", the gold target is "Outstanding Paper Award". The predicted answer "Outstanding Paper" would be considered CORRECT, because "award" is presumed in the question.
    - For the question "What is the height of Jason Wei in meters?", the gold target is "1.73 m". The predicted answer "1.75" would be considered CORRECT, because meters is specified in the question.
    - For the question "What is the name of Barack Obama's wife?", the gold target is "Michelle Obama". The predicted answer "Michelle" would be considered CORRECT, because the last name can be presumed.
- Do not punish for typos in people's name if it's clearly the same name. 
    - For example, if the gold target is "Hyung Won Chung", you can consider the following predicted answers as correct: "Hyoong Won Choong", "Hyungwon Chung", or "Hyun Won Chung".

Question: {question}
Gold target: {ground_truth}
Predicted answer: {solution_str}

Simply reply with either CORRECT, INCORRECT. Don't apologize or correct yourself if there was a mistake; we are just trying to grade the answer.
\end{lstlisting}
\end{figure*}

\section{Training Prompt}
\label{sec:training-prompt}

The following prompt is used to train the Deliberative Searcher models to integrate search operations with confidence calibration during chain-of-thought generation:

\begin{figure*}[!ht]
\lstset{
  frame=single,
  framerule=1pt,
  rulecolor=\color{gray},
  backgroundcolor=\color{gray!5},
  basicstyle=\footnotesize\ttfamily,
  breaklines=true,
  caption={Training prompt for Deliberative Searcher models.},
  label=lst:training-prompt,
  captionpos=t
}

\begin{lstlisting}
You are an AI assistant. To answer questions, use the following XML-like tags to structure your reasoning process, interact with a search engine, and formulate your answer.

<think>Your reasoning here</think>: Enclose your step-by-step reasoning. Use this to analyze information, plan next steps (like searching or reading), and process document contents.
<search>Your search query</search>: Issue a search query to the simulated search engine.
<result>...</result>: (Provided by the system) Contains the search results, typically a list of documents with titles and abstracts.
<read>X</read>: Command to retrieve the full text of document X (where X is the document number from search results). This command **must** be placed immediately after the </think> block in which you decided to read document X.
<document X>Full text of document X</document X>: (Provided by the system after a <read X> command) The full content of the specified document.
<answer>Summary leading to final answer, with citations</answer>: Present the answer to the user's question. Cite information from read documents using [citation:X] immediately after the information derived from it.
<CONFIDENCE:y>: An integer from 0 (very uncertain) to 10 (very confident).
    * For each <search> or <read> operation, assess its utility, relevance, and the clarity of the information obtained, expressing this assessment with an <CONFIDENCE:x> tag. Crucially, this tag MUST be outputted after processing the current operation's results and strictly before initiating any subsequent <search> or <read> operation.
    * Append this at the very end of your <answer> block (e.g., <answer>... <CONFIDENCE:x></answer>) to state your overall confidence in the final answer's accuracy and completeness.

Here's a minimal example of how these tags flow in an interaction:
<think>Initial analysis of the question. <CONFIDENCE:a> So I need to search for X.</think>
<search>X related query</search>
<result>
[
     {"id": 1, "title": "Title of Document 1", "abstract": "Abstract of Document 1..." },
     {"id": 2, "title": "Title of Document 2", "abstract": "Abstract of Document 2..." },
     ...
]
</result>
<think>Evaluated search results. Document 1 seems relevant.<CONFIDENCE:a></think>
<read>n</read>
<document n>
This is the full text content of Document n. It contains key information Y.
</document n>
<think>Processed Document n. Key information Y was found. <CONFIDENCE:b></think>
... (more read if needed)
<think>Based on those information, I can know Y and Z <CONFIDENCE:c></think>
... (more search and read if needed)
<answer>Begin by presenting key information derived from your readings, for instance, "Source X states that relevant fact or finding from source X [citation:X]."
You can then add further relevant details or related findings, e.g., "This is complemented by information from Source Y, which indicates another relevant detail from source Y [citation:Y]. "
Continue to build the necessary factual foundation by summarizing other useful points from your readings, ensuring each is cited, e.g., "Additionally, key point from source Z [citation:Z] is important to consider."
Then, transition to the direct answer, for example, "Based on this collective information," or "Therefore,".
Provide the synthesized answer to the user's question, drawing from the previously summarized points, e.g., "it can be concluded that your synthesized answer, which might reference insights from [citation:X] and [citation:Y]."
If needed, you can add further clarification or nuances to your answer here.<CONFIDENCE:d></answer>
\end{lstlisting}
\end{figure*}

\section{Document Processing}

During the search process, documents are processed to generate abstracts for initial presentation in search results. The system truncates document content to create concise abstracts (first 50 characters) while preserving document titles and IDs for subsequent full-text retrieval. This two-stage retrieval design enables the model to first assess relevance from abstracts before committing to reading full documents, creating explicit decision points that generate richer training signals for the reinforcement learning algorithm. The implementation assigns unique IDs to documents and maintains the mapping between search results and full document content throughout the search trajectory.

\section{Impact of Constrained Reinforcement Learning}
\label{sec:impact of constrained RL}

\begin{figure}[t]
    \centering
    
    \begin{subfigure}[b]{0.48\columnwidth}
        \centering
        \includegraphics[width=\linewidth]{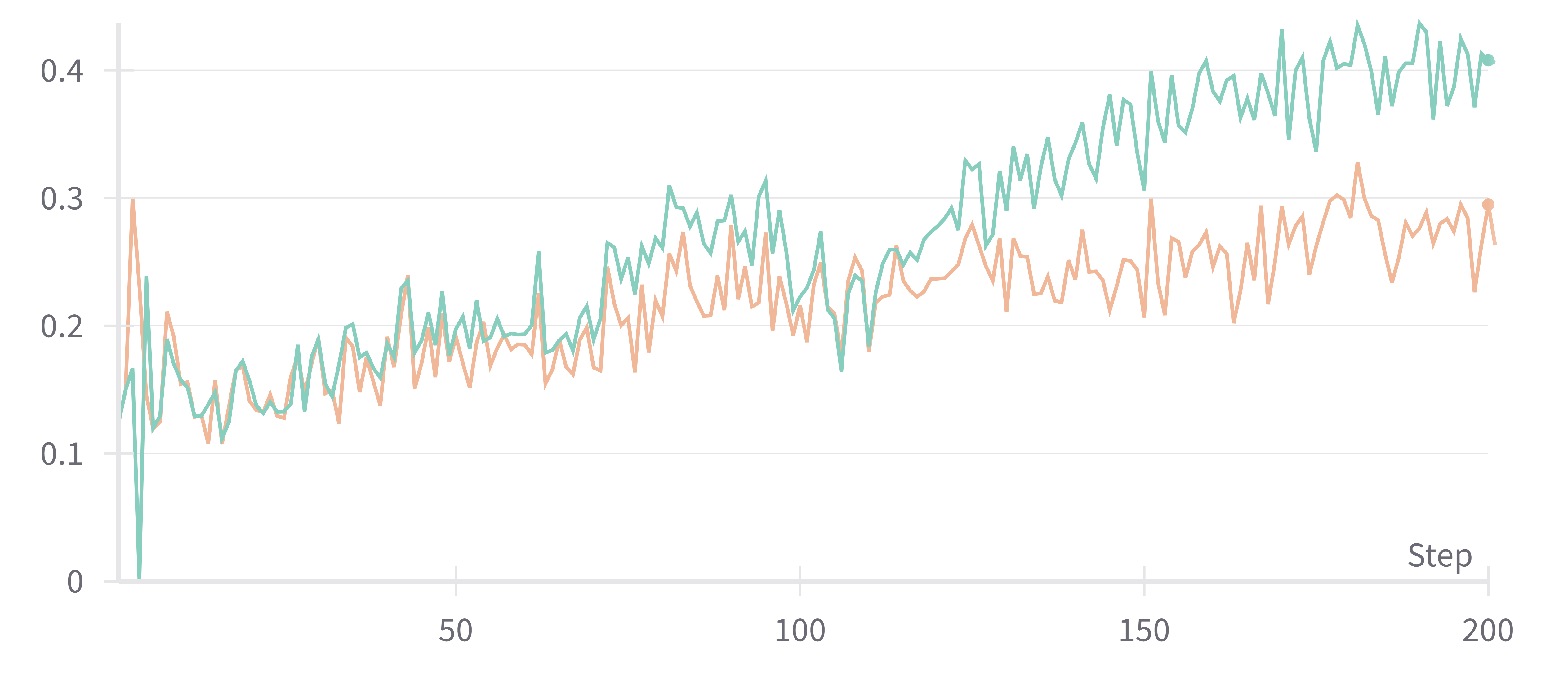}
        \caption{Correctness}
        \label{fig:image1}
    \end{subfigure}
    \hfill 
    \begin{subfigure}[b]{0.48\columnwidth}
        \centering
        \includegraphics[width=\linewidth]{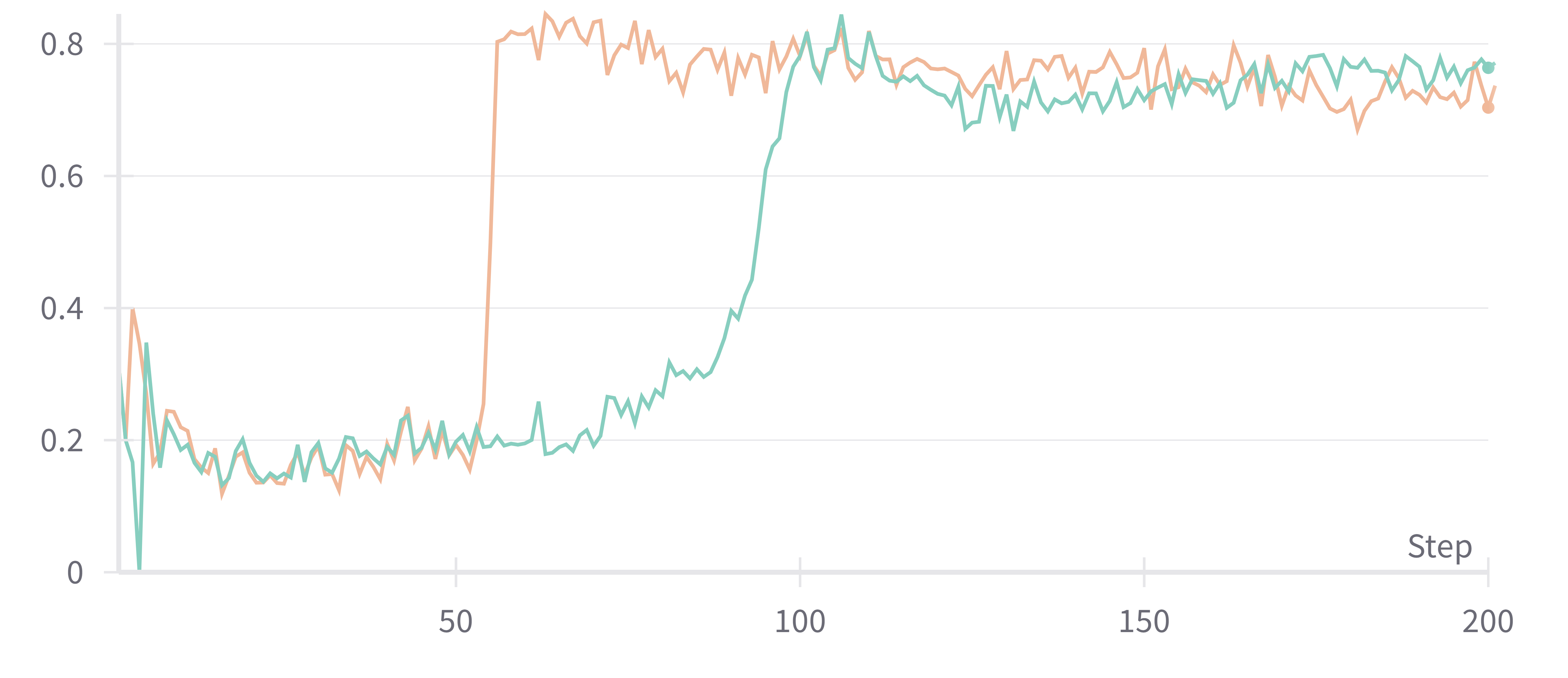}
        \caption{Reliability}
        \label{fig:image2}
    \end{subfigure}
    
    
    \begin{subfigure}[b]{0.48\columnwidth}
        \centering
        \includegraphics[width=\linewidth]{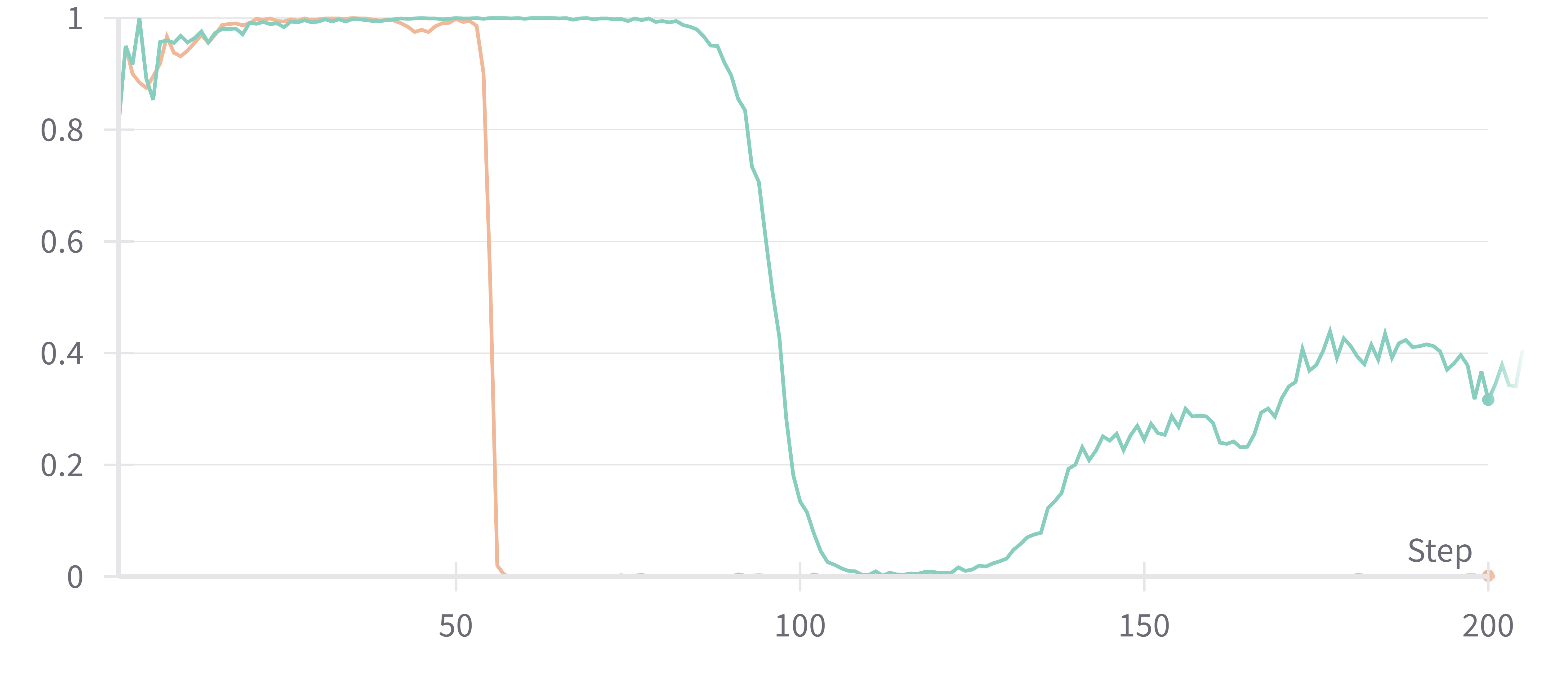}
        \caption{Certainty}
        \label{fig:image3}
    \end{subfigure}
    \hfill 
    \begin{subfigure}[b]{0.48\columnwidth}
        \centering
        \includegraphics[width=\linewidth]{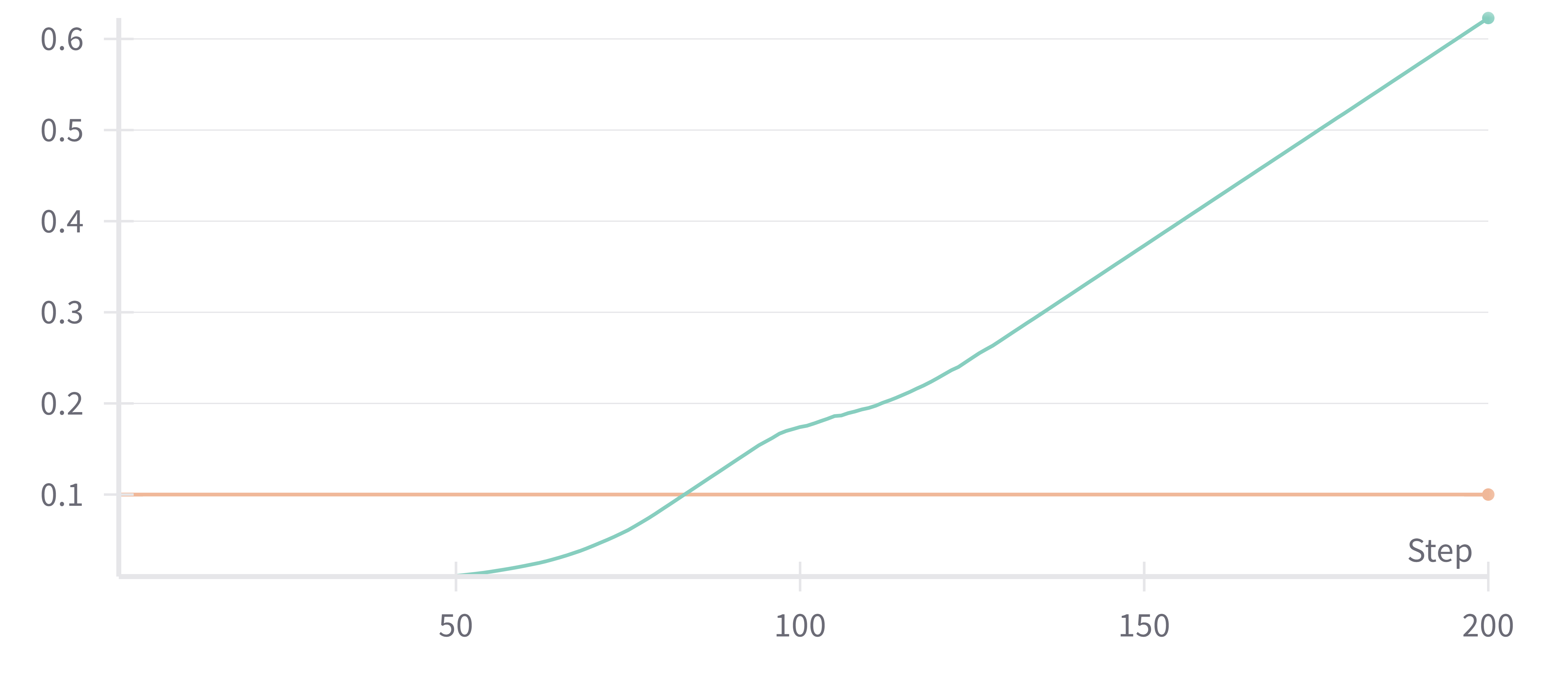}
        \caption{Lagrange Multiplier $\lambda$}
        \label{fig:image4}
    \end{subfigure}
    
    \caption{Ablation study of the constrained reinforcement learning algorithm. \textcolor[RGB]{135,206,191}{Green} lines represent our adaptive method while \textcolor[RGB]{240,184,153}{orange} lines show the fixed-weight baseline.}
    \label{fig:ablation}
\end{figure}

To validate the necessity of our constrained RL approach, we conduct an ablation study comparing our adaptive Lagrange multiplier scheme against a fixed-weight baseline. Importantly, the fixed-$\lambda$ baseline ($\lambda = 0.1$) is mathematically equivalent to training with a composite reward function $r_{final} = r_{acc} + \lambda \cdot r_{reliab}$ where the weights are fixed throughout training. This comparison directly addresses whether the constrained optimization machinery is necessary, or if a simpler weighted reward could achieve comparable results.

Both methods use identical GRPO training configurations, with the key difference being the reliability weight: our method adaptively updates $\lambda$ using learning rate $\eta = 0.1$ and reliability threshold $a = 0.9$, while the baseline maintains fixed $\lambda = 0.1$ throughout training.

As shown in Figure~\ref{fig:ablation}, the fixed-weight baseline catastrophically fails. The certainty plot (c) reveals that the baseline collapses to always predicting low confidence around step 100, representing a degenerate solution where the model learns to trivially satisfy the reliability objective by never expressing certainty. In contrast, our adaptive method maintains meaningful confidence distinctions throughout training.

The training dynamics in panels (b) and (c) reflect expected optimization behavior: around step 100, as accuracy plateaus, the increasing constraint (panel d) forces the model to reconsider its calibration strategy. While the model explores various strategies during this phase, adaptive constraints successfully guide it away from degenerate solutions, ultimately achieving balanced performance between accuracy and reliability.

The key advantage of our adaptive approach is \textbf{transferability}: while there may exist fixed weights that work well for a specific model, finding them requires extensive hyperparameter search, and optimal values often fail when transferred to different architectures. Our constraint-based formulation specifies the desired reliability level directly (e.g., $\geq 90\%$ reliability), allowing the optimizer to automatically find appropriate $\lambda$ values. This eliminates expensive hyperparameter tuning when scaling to new architectures, as demonstrated by consistent performance across our 7B, 72B, and DeepSeek-70B variants without any architecture-specific adjustments.